\def\year{2018}\relax
\documentclass[letterpaper]{article} 
\usepackage{aaai18}  
\usepackage{times}  
\usepackage{helvet}  
\usepackage{courier}  
\usepackage{url}  
\usepackage{graphicx}  
\usepackage{hyperref}
\usepackage{amsmath}
\usepackage{amsfonts}
\usepackage{algorithm}
\usepackage{algpseudocode}
\usepackage{subcaption}
\usepackage{float}
\algnewcommand\algorithmicinput{\textbf{Input:}}
\algnewcommand\INPUT{\item[\algorithmicinput]}
\begin{document}
\title{Shared Learning: Enhancing Reinforcement in $Q$-Ensembles}
\author{Rakesh R Menon \\
College of Information and Computer Sciences\\
University of Massachusetts Amherst\\
\href{mailto:rrmenon@cs.umass.edu}{rrmenon@cs.umass.edu}
\And
Balaraman Ravindran\\
Department of Computer Science and Engineering\\
Robert Bosch Centre for Data Science and AI\\
Indian Institute of Technology Madras\\
\href{mailto:ravi@cse.iitm.ac.in}{ravi@cse.iitm.ac.in}
}
\maketitle
\begin{abstract}
Deep Reinforcement Learning has been able to achieve amazing successes in a variety of domains from video games to continuous control by trying to maximize the cumulative reward. However, most of these successes rely on algorithms that require a large amount of data to train in order to obtain results on par with human-level performance. This is not feasible if we are to deploy these systems on real world tasks and hence there has been an increased thrust in exploring data efficient algorithms. To this end, we propose the Shared Learning framework aimed at making $Q$-ensemble algorithms data-efficient. For achieving this, we look into some principles of transfer learning which aim to study the benefits of information exchange across tasks in reinforcement learning and adapt transfer to learning our value function estimates in a novel manner. In this paper, we consider the special case of transfer between the value function estimates in the $Q$-ensemble architecture of BootstrappedDQN. We further empirically demonstrate how our proposed framework can help in speeding up the learning process in $Q$-ensembles with minimum computational overhead on a suite of Atari 2600 Games.
\end{abstract}

\section{Introduction}

Reinforcement Learning (RL) deals with learning how to act in an environment by trying to maximize the cumulative payoff \cite{sutton1998reinforcement}. However, the early approaches in RL did not scale well to environments with large state spaces. Recently, Deep Reinforcement Learning (DRL) has gained a great deal of interest because of its ability to map high-dimensional observations to actions using a neural network function approximator \cite{mnih2015human,wang2016dueling,schaul2015prioritized,mnih2016asynchronous}. With the development of many algorithms, researchers have been able to show the effectiveness of deep reinforcement learning in trying to solve problems in complex domains. However, most of these architectures require a large amount of training data in order get near human-level performance. This problem of \textit{data inefficiency} has been well established previously in \cite{lake2016building}. \\
\\
Prior attempts at improving data efficiency in reinforcement learning, involved the use of an Experience Replay mechanism \cite{lin1992self} which allowed the agent to replay trajectories from a memory buffer to learn more effectively from each sample trajectory. This idea was first introduced in deep Q-learning by \cite{mnih2015human}. Later, \cite{wang2016sample} was able to use the idea of Experience Replay to create a sample efficient actor-critic along with some other modifications. Hindsight Experience Replay \cite{2017arXiv170701495A} tries to add samples to the replay memory which help the agent to learn about both the desired and undesired states in the environment more efficiently. Another line of work involves auxiliary tasks that learn how to control the environment on the high dimensional visual observations \cite{jaderberg2016reinforcement,2016arXiv160905521L}. More recently, combining model-based methods and model-free methods have been shown to be make learning data-efficient \cite{2017arXiv170706203W}. We aim to develop an agent that can learn more efficiently from each sample in the Experience Replay.\\
\\
The framework we propose builds upon some ideas from transfer in reinforcement learning literature \cite{taylor2009transfer}. In particular, we have focused on transfer through \textit{action advice} \cite{taylor2014reinforcement} to perform online transfer. While online transfer has been done before in \cite{zhan2015online}, our framework differs from this algorithm in the sense that the \textit{action advice} happens while learning from samples collected in the Experience Replay. A $Q$-ensemble is an suitable architecture to perform online transfer since we can take advantage of the independent value function estimates within the architecture. The BootstrappedDQN architecture \cite{osband2016deep} is an example of one such $Q$-ensemble algorithm that learns to explore complex environments more efficiently by planning over several timesteps. Experimentally, BootstrappedDQN was shown to perform better than Double DQN \cite{van2016deep} on most Atari 2600 games. \cite{2017arXiv170601502C} proposed 3 algorithms, Ensemble Voting, UCB exploration and UCB+InfoGain exploration, inspired by concepts from bayesian reinforcement learning and bandit algorithms. These algorithms were shown to perform better than BootstrappedDQN and Double DQN on many Atari 2600 games. In this paper, we present a new framework, \textit{Shared Learning}, that learns to share knowledge between the value function estimates of BootstrappedDQN and Ensemble Voting which allows for data-efficiency and then show how our framework can be extended to any $Q$-ensemble algorithm.\\
\\
The rest of the paper is structured as follows: we will look at some related works in data-efficient reinforcement learning, $Q$-ensembles and transfer learning. Following which we touch upon some necessary background before moving on to the main section on Shared Learning. We first motivate and analyze how our framework helps through toy Markov Decision Process(MDP) chains. Further, we go on to show the efficacy of our proposed framework on the Arcade Learning Environment \cite{bellemare2013arcade} Atari 2600 games.

\section{Related Work}

\subsection{Data Efficiency}

With the evolution of deep learning and RL, many algorithms were developed which could solve many complex problems in domains like robotics and games. However, most of these algorithms suffer from some fundamental issues that were highlighted in \cite{lake2016building}. Some of these fundamental issues include data inefficiency, brittle nature of learned policies and the inability to adapt to new tasks. Here, we focus on some works that try to tackle the problem of data efficiency. The first work on deep Q-learning \cite{mnih2015human} introduced the idea of Experience Replay \cite{lin1992self} to improve data-efficiency . By replaying the trajectories visited by the agent, the agent was able to decorrelate samples for training and also learn more robustly from each sample. \cite{wang2016sample} was able to apply the idea of Experience Replay along with some other modifications into the on-policy actor-critic and make the whole architecture sample efficient. Recently, \cite{2017arXiv170701495A} proposed Hindsight Experience Replay, a method allowing the agent to learn as much information about the undesired outcomes as it would of the desired outcomes. To do this they take as input, the observation and the goal state that is to be achieved (experiments were done on environments where the goal state was known) and train the neural network to act in order to get to the goal state. They further add samples in the Experience Replay that reward the agent upon reaching different states visited in the agents trajectories. This allows for more information to be learned about the environment.\\
\\
\cite{jaderberg2016reinforcement} introduced a method of learning multiple unsupervised auxiliary tasks on the visual data stream that allow the agent to learn to control and predict different aspects of the environment. The proposed agent, UNREAL, significantly outperformed previous baselines based on the on-line A3C \cite{mnih2016asynchronous} architecture. The algorithm also learned robust policies and was able to learn with much less data when compared to A3C. \cite{2016arXiv160905521L} tries to predict the actions that are required to be taken by the agent along with predictions of some game features which lead to an increase in performance on VizDoom \cite{2016arXiv160502097K}. \cite{2016arXiv160602396K} also tries to reproduce the current state of the environment as an auxiliary task along with the successor representation of the value function.\\
\\
Model-based methods are known to be one of the best methods for data-efficient learning because such algorithms can perform planning. However, such algorithms are hinged on the fact that they have access to a model of the environment. This is not such an easy task in the case of deep reinforcement learning. \cite{2017arXiv170706203W} has been able to achieve some success by combining model-based and model-free methods to get a data-efficient learning algorithm. Their algorithm makes use of an imperfect model of the environment to create imagination-based trajectories and the final policy and value function are provided as a combination of the model-based and the model-free algorithm.

\subsection{Exploration Strategies}

Exploration strategies in reinforcement learning have been able to achieve near-optimal guarantees for many small domains like Markov Decision Process chains and puddle world in the past \cite{brafman2002r,kearns1999efficient}. However, these near-optimal algorithms were not able to scale well to large domains that deep reinforcement learning aims to solve. Most of the recently proposed exploration strategies, have made use of pseudo counts for state visitations \cite{bellemare2016unifying,2017arXiv170301310O}, hashing \cite{DBLP:journals/corr/TangHFSCDSTA16}, exploration bonuses\cite{DBLP:journals/corr/StadieLA15} and intrinsic motivation \cite{houthooft2016vime}. However, there was a need for strategies that can perform deep exploration over multiple time-steps.\\
\\
The $Q$-ensemble architecture was first introduced in DRL by \cite{osband2016deep} through the BootstrappedDQN architecture to perform deep exploration along with some level of planning. The architecture derives most of its inspiration from Posterior Sampling for Reinforcement Learning (PSRL) \cite{osband2013more}. BootstrappedDQN maintains multiple $Q$-value estimates and was able to perform more efficient exploration than Double DQN \cite{van2016deep} on majority of the Atari Games. Recently, \cite{2017arXiv170601502C} modified BootstrappedDQN to further improve exploration by adapting the UCB algorithm \cite{auer2002finite} to the $Q$-ensemble architecture and further using the disagreement among the value function estimates in order to provide a reward bonus signal. Additionally, the paper also proposed an exploitation $Q$-ensemble algorithm called Ensemble Voting where the agent follows a policy which is the majority vote of the heads.

\subsection{Transfer in DRL}

Transfer for reinforcement learning has been studied to a great extent in the past \cite{taylor2009transfer}. While several methods for performing transfer have been proposed in the past, here we study \textit{action advice} in reinforcement learning. In \textit{action advice}, we have a teacher advice the student what kind of action needs to be taken at each step. A2T \cite{rajendran2017attend} tries to provide action advice from multiple trained experts to a new agent to try and capture different skills from each teacher through positive policy transfer. Other methods  like \cite{parisotto2016actor,rusu2016policy} have involved knowledge transfer from one (or multiple) source tasks to a target task in order to speed up the learning process by learning from transitions of the expert agents. These methods were also shown to be able to learn more general representations of the environment on Atari 2600 games. However, these algorithms depend on an agent that has been trained completely on a task(or a teacher) to perform the knowledge transfer to a fresh agent(or a student). \cite{zhan2015online} is one of the first papers to present online transfer between agents that are still learning how to act in the environment. The paper further goes on to show that classical transfer is a specific case of online transfer and prove the convergence of Q-learning and SARSA using online transfer.
\section{Background}
\subsection{RL Notations}
A common approach to solving an RL problem is through value functions that indicate the expected reward that can be obtained from each state. In this paper, we only concern ourselves with state-action value functions $Q(s,a)$ which is the return that an agent is expected to receive from a particular state $s$ upon taking an action $a$. One common approach for learning the value functions is an off-policy TD-algorithm called Q-learning \cite{watkins1992q}. The optimal policy can be achieved by behaving greedily with respect to the learned state-action value function in each state. The update rule for the Q-learning is as follows,
\begin{flalign}\label{eqwat}
     \nonumber Q_{t+1}(s_t,a_t) = Q_t(s_t,a_t) + \alpha( &r_t+\gamma \text{max}_{a}Q_t (s_{t+1},a)\\ & - Q_t(s_t,a_t))
\end{flalign}
Here, $s_t$ is the state, $a_t$ is the action taken and $r_t$ is the reward obtained by taking that action at time $t$. From here on, we will refer to $Q(s,a)$ as value function.

\subsection{Deep Q-Networks (DQN)}
In environments with large state spaces, it is not possible to learn values for every possible state-action pair. The need for generalizing from experience of a small subset of the state space to give useful approximations of the value function becomes a key issue \cite{sutton1998reinforcement}. Neural networks, while attractive as potential value function approximators, were known to be unstable or even to diverge on reinforcement learning problems. Through some seminal work on representation learning using deep neural networks \cite{lecun2015deep} were able to create new methods for learning algorithms from the high dimensional visual observations. \cite{mnih2015human} successfully used deep neural networks in order to carry out Q-learning using the high dimensional Atari 2600 \cite{bellemare2013arcade} screen observation as input. Additionally, \cite{mnih2015human} also overcomes the problem of stability in learning using two crucial ideas: \textit{replay memory} and \textit{target networks}. The parametrized value function is trained using the expected squared TD-error as the loss signal given to the neural network.
\begin{flalign}
         \nonumber L_i(\theta_i) = \mathbb{E}_{s,a,r,s'}[(&(r+ \gamma max_{a'}Q(s',a';\theta_i^{-})\\ & - Q(s,a;\theta_i))^2]
\end{flalign}
Here, $\theta_i$ represents the weights of \textit{online network} and $\theta_i^{-}$ refers to the \textit{target network} weights. The \textit{target network} ensures that the learning happens in a stable manner while the \textit{replay memory} ensures that network can learn from independent identically distributed(i.i.d.) samples hence ensures stable neural network training.\\
\\
The max operator in Q-learning has been shown to produce an overestimation error in the value function \cite{hasselt2010double}. To overcome this overestimation error in DQNs, \cite{van2016deep} introduced Double Deep Q-Networks (DDQN), a low computational overhead algorithm, in which the greedy policy is evaluated using the online network and the value is given by the target network for the learning update. Thus the loss signal for the DDQN becomes:
\begin{flalign}
         \nonumber L_i(\theta_i) = \mathbb{E}_{s,a,r,s'}[ & (r+ \gamma Q(s', \text{argmax}_{a'}Q(s',a';\theta_i);\theta_i^{-}) &&\\
         & - Q(s,a;\theta_i))^2]
         \label{eq:ddqn}
\end{flalign}
\subsection{BootstrappedDQN}
\begin{figure}[H]
\centering
\includegraphics[scale=0.4]{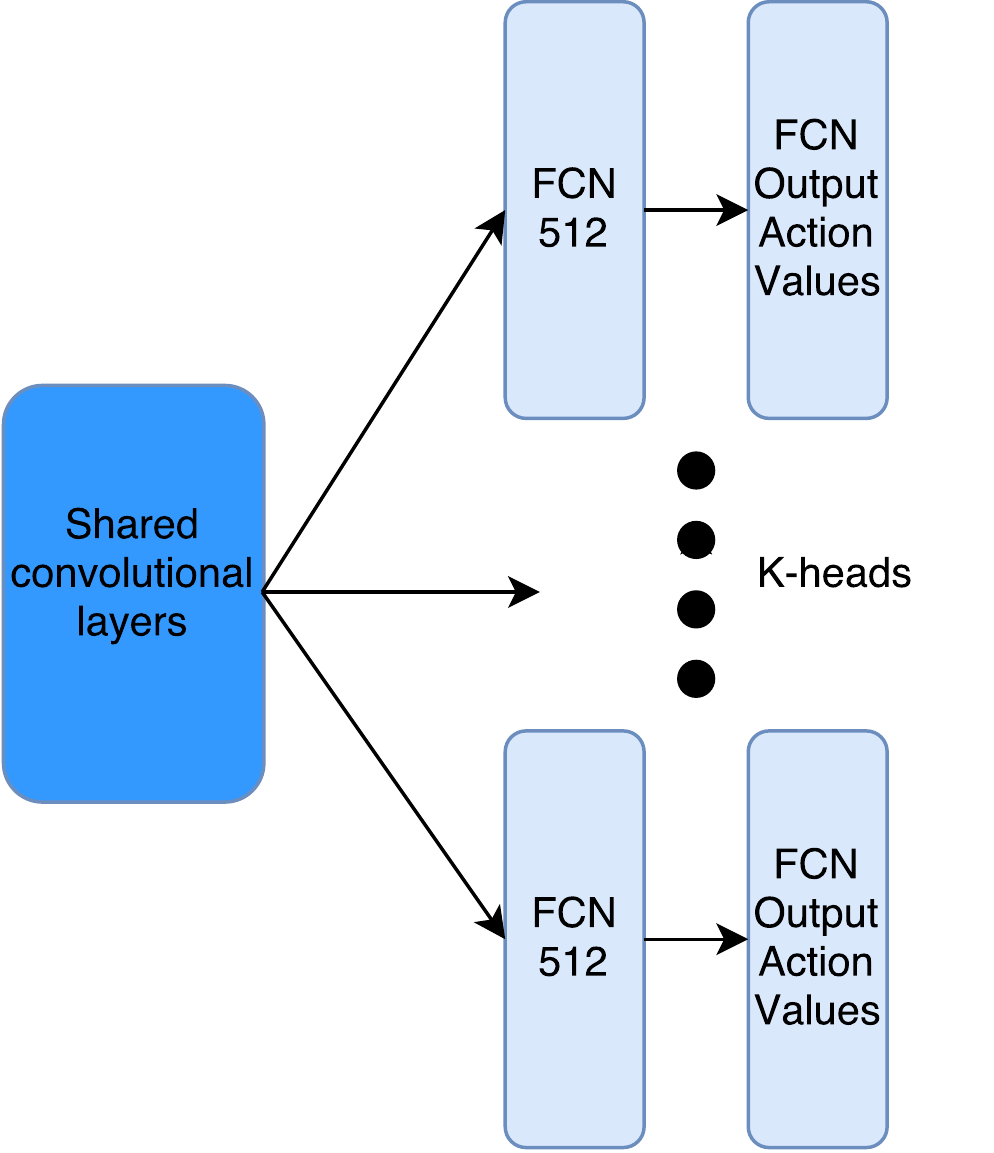}
\caption{BootstrappedDQN architecture}
\label{fig:bootdqn}
\end{figure}
\begin{figure*}[t!]
	\centering
    \begin{subfigure}[b]{0.33\textwidth}
        \includegraphics[width=\textwidth]{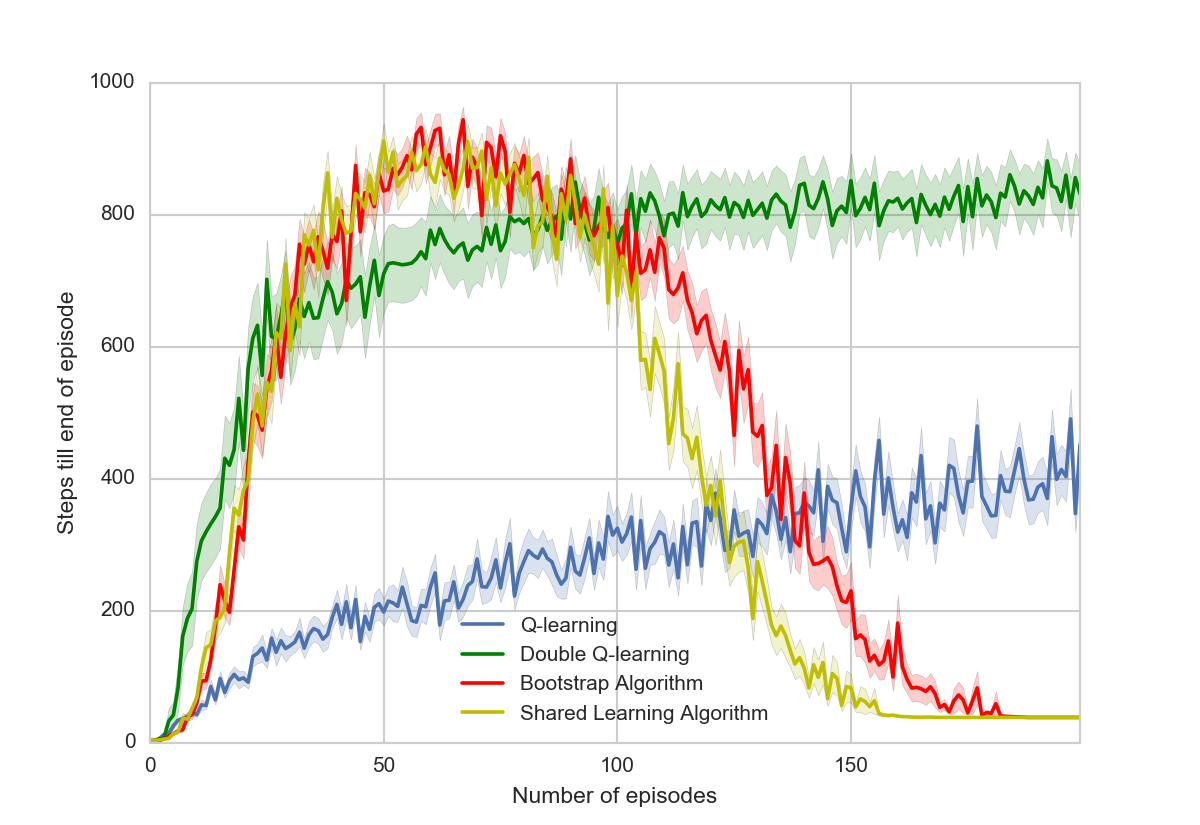}
        \caption{40-state chain MDP}
        \label{fig:Seaquest}
    \end{subfigure}
    \centering
    \begin{subfigure}[b]{0.33\textwidth}
        \includegraphics[width=\textwidth]{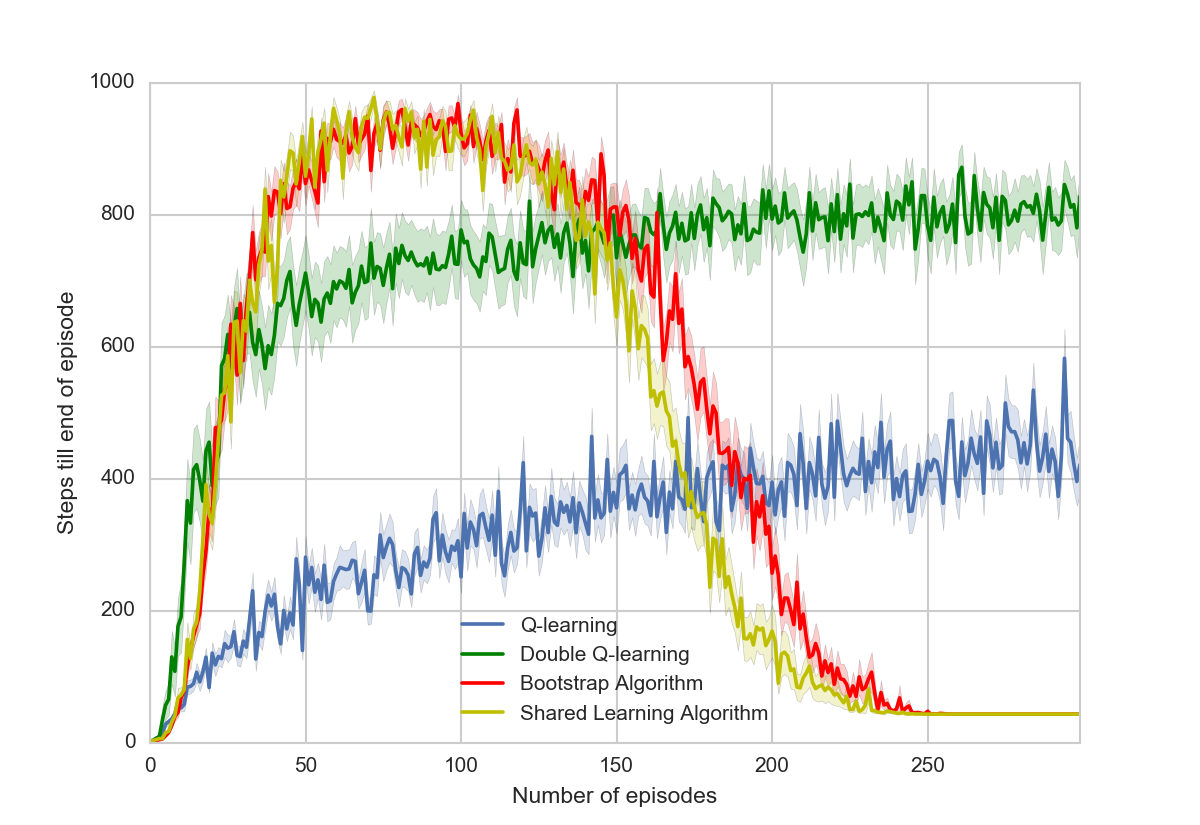}
        \caption{45-state chain MDP}
        \label{fig:Seaquest}
    \end{subfigure}
    \centering
    \begin{subfigure}[b]{0.33\textwidth}
        \includegraphics[width=\textwidth]{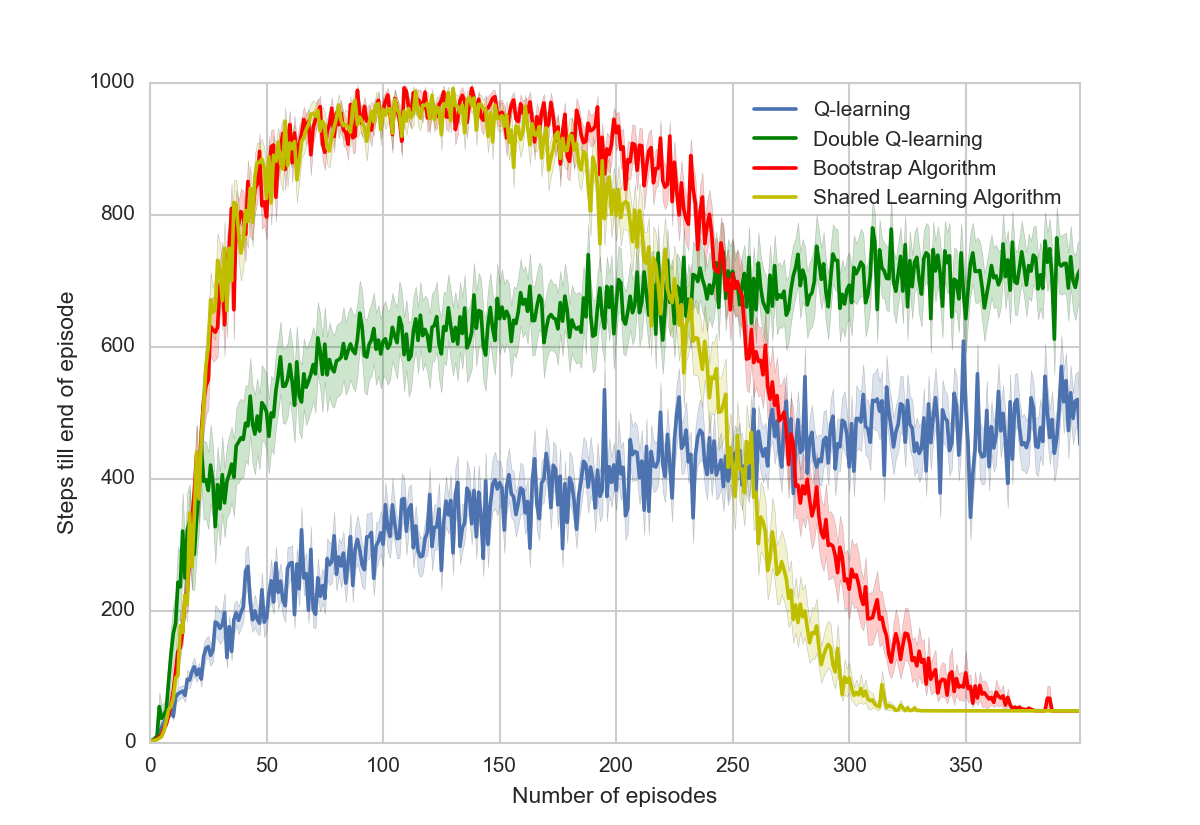}
        \caption{50-state chain MDP}
        \label{fig:Seaquest}
    \end{subfigure}
    \caption{Comparison of number of steps taken by Shared Learning-Bootstrap, Bootstrap, Q-learning and Double Q-learning algorithm during each run (averaged over 50 runs). When the algorithm has converged there is a zero variance line corresponding to the fastest path to the goal state which takes $n-2$ steps in an $n$-state MDP chain.}
   	\label{fig:mdpsteps}
\end{figure*}
BootstrappedDQN \cite{osband2016deep} introduces a novel exploration strategy that performs deep exploration in environments with large state spaces. The idea mainly draws inspiration from two prior works on PSRL \cite{osband2013more} and RLSVI \cite{DBLP:conf/icml/OsbandRW16}. The $Q$-ensemble architecture maintains multiple parametrized value function estimates or ``heads". BootstrappedDQN depends on the independent initializations of the heads and the fact that each head is trained with different set of samples. However, it was shown that training each head with different samples was not as much an important criteria as the independent initializations.\\
\\
At  the  start  of  every  episode,  BootstrappedDQN  samples  a  single  value  function  estimate  at  random according to a uniform distribution.  The agent then follows the  greedy  policy  with  respect  to  the  selected  estimate until  the  end  of  the  episode. The  authors  propose  that this is an adaptation of the Thompson sampling heuristic to RL that allows for temporally extended (or deep) exploration.\\
\\
BootstrappedDQN   is   implemented   by   adding   multiple heads which branch out from the output of the convolutional layers as  shown  in  Figure  \ref{fig:bootdqn}.   Suppose  there  are $K$ heads  in  this network.    The  outputs  from  each  head  represent  different independent  estimates  of  the  action-value  function. Let $Q_k(s,a;\theta_i)$ be  the  value  estimate  and $Q_k(s,a;\theta_i^-)$ be  the target value estimate of the $k^{th}$ head.  The loss signal for the kth head is given as follows,
\begin{flalign}
         \nonumber L_i(\theta_i) = \mathbb{E}_{s,a,r,s'}[ & (r+ \gamma Q_k(s', \text{argmax}_{a'}Q_k(s',a';\theta_i);\theta_i^{-}) &&\\
         & - Q_k(s,a;\theta_i))^2]
         \label{eq:bootdqn}
\end{flalign}
Each of the $K$ heads is updated this way and the gradients are aggregated and normalized at the convolutional layers.

\section{Shared Learning}
\subsection{A motivating example}
\begin{figure}[h]
	\centering
	\includegraphics[scale=0.38]{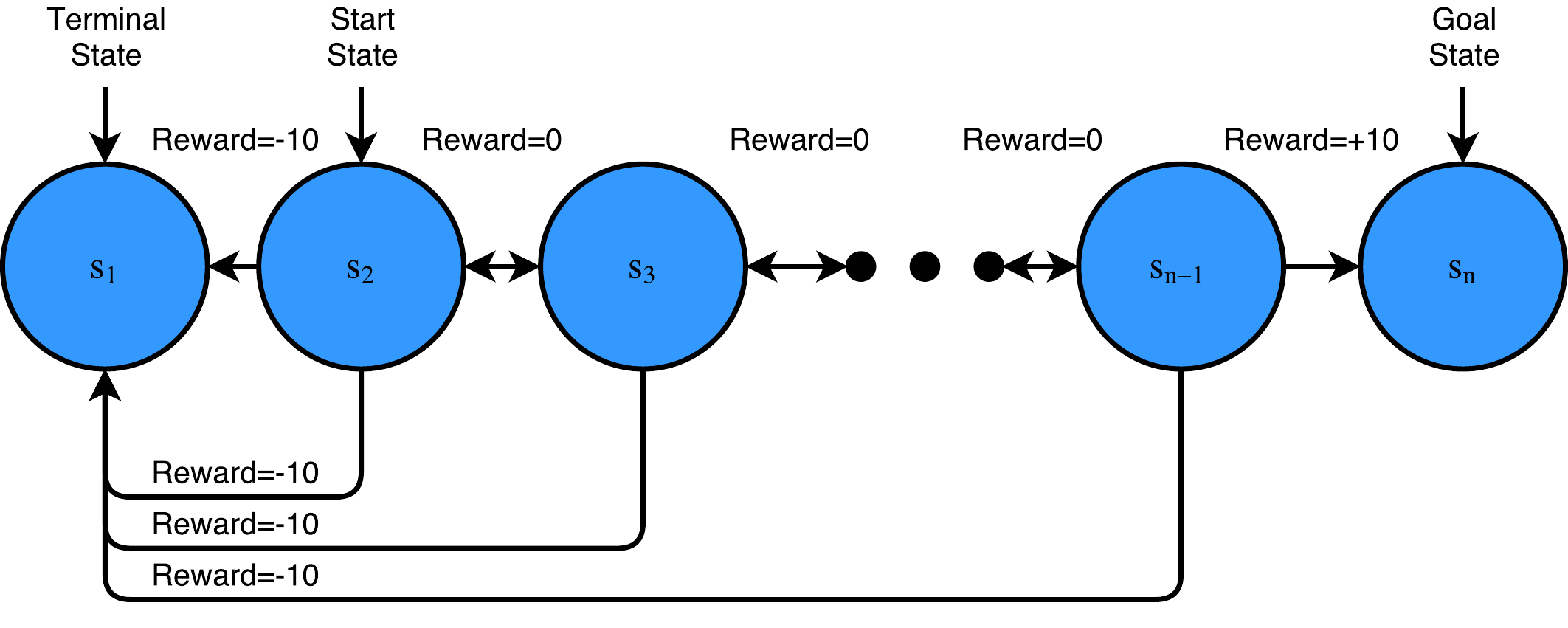}
    \caption{MDP chain with $n$-states}
    \label{fig:mdpchain}
\end{figure}

Consider the task of solving an MDP chain environment as in Figure \ref{fig:mdpchain} with state space $\mathcal{S}$ = $\{s_1,s_2,...,s_n\}$ and action space $\mathcal{A}$ = \{\textit{Jump to $s_1$}, \textit{Right}, \textit{Left}, \textit{No-op}\}. In this task, the agent begins at state $s_2$ and has to reach state $s_n$ to get a reward of +10 and if it goes to state $s_1$, it gets a reward of -10. The episode terminates when either state $s_1$ or $s_n$ is reached.\\
\\
This is an environment that good exploration algorithms can solve very fast. Algorithms like Q-learning and Double Q-learning \cite{hasselt2010double} find it very hard to solve such an environment. As shown in the graph in Figure \ref{fig:mdpsteps}, a tabular adaptation of BootstrappedDQN, which we call the \textit{Bootstrap} algorithm here, is able to solve larger chains when compared to Q-learning \cite{watkins1992q} and Double Q-learning. However, the algorithm takes a long time to converge to the goal state because all the value function estimates have to learn from their independent experiences. So, we need a method to transfer the knowledge among other value function estimates. This where \textit{Shared Learning} steps in, our framework for $Q$-ensemble algorithms which is able to allow the ensemble estimates to learn from each other through action advice in the learning update. The results for Shared Learning, in Figure \ref{fig:mdpsteps}, indicate that our framework converges to the goal state solution faster than Bootstrap.
\subsection{Knowledge Sharing in Learning Update}
The motivation for Shared Learning is to allow the different value function estimates to learn from each other so that the complete agent can solve a task much faster and more efficiently. To this end, knowledge must be transferred from an apparent expert to the other estimates. Transfer learning literature suggests that in order to make sure that other estimates are able to replicate the rewarding trajectories of the expert, we should perform action advice. Instead, we introduce a novel method of trying to influence an estimate to attempt to go along the direction of the chosen estimate through a minor modification in the learning update of BootstrappedDQN as given in Equation \ref{eq:slp_arbit}.\\
\begin{flalign}
         \nonumber L_i(\theta_i) =  \mathbb{E}_{s,a,r,s'}[ &(r+ \gamma Q_k(s', \text{argmax}_{a'}Q_{m}(s',a';\theta_i);\theta_i^{-}) &&\\
         & - Q_k(s,a;\theta_i))^2]
         \label{eq:slp_arbit}
\end{flalign}
where $m \neq k$. Since our requirement is to follow directions along maximum reward, we can choose the estimate that is able to give the highest(best) expected reward to perform the knowledge sharing. So, we modify the above equation to give the following update rule,
\begin{flalign}
         \nonumber L_i(\theta_i) =  \mathbb{E}_{s,a,r,s'}[ &(r+ \gamma Q_k(s', \text{argmax}_{a'}Q_{best}(s',a';\theta_i);\theta_i^{-}) &&\\
         & - Q_k(s,a;\theta_i))^2]
         \label{eq:slp}
\end{flalign}
While \cite{zhan2015online} tried to show how action advice during learning can help in the overall training process we have shown, in a novel manner, how  to use \textbf{action advice in the learning update} to perform faster learning.
\subsection{Robust Target Estimates}
\begin{figure}[h!]
\centering
\includegraphics[width=0.45\textwidth]{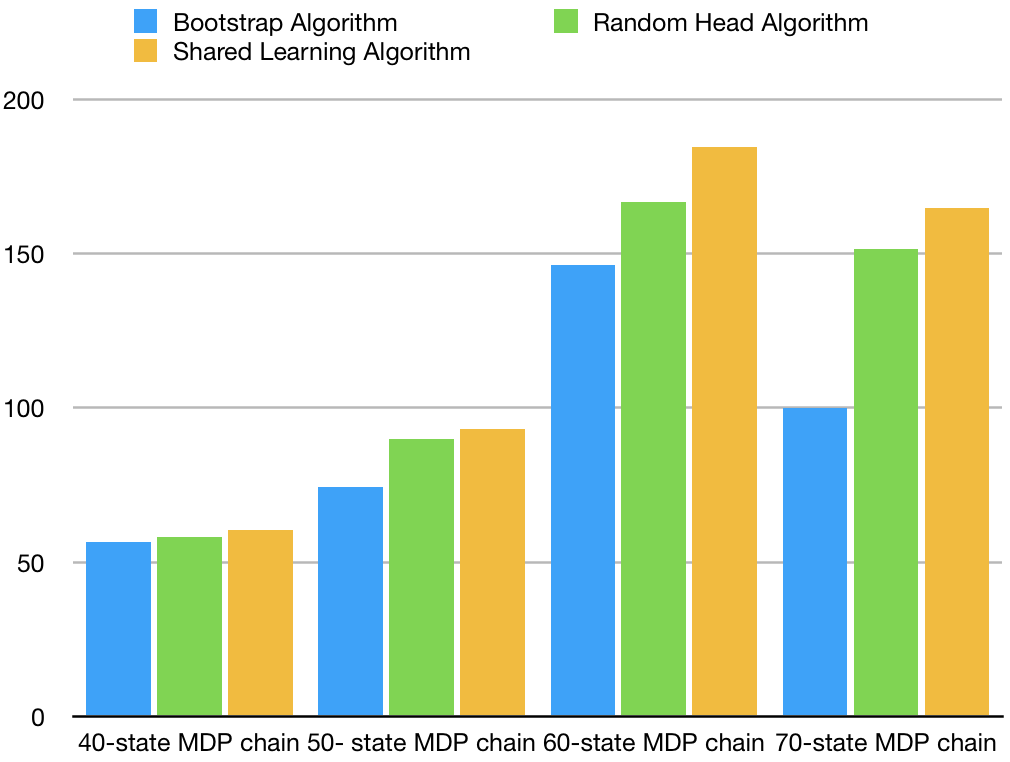}
\caption{Comparison of the number of times the goal state is reached by Shared Learning, Bootstrap and Random Head algorithm on the MDP chain environment. }
\label{fig:mdpreach}
\end{figure}

While proposing to use Double Q-learning in DQNs in \cite{van2016deep}, the author(s) proposed a low computational overhead modification which was shown to reduce the overestimation errors in DQNs by using the online network to provide the greedy policy while using the target network for the value estimate. But, the target network, being a previous iteration of the online network, is coupled with the online network. This reduces the capability of the Double Q-learning update and has been reported in \cite{van2016deep}.\\
\\
However, through Shared Learning we can ensure that most of the time the estimate that is providing the greedy action ($Q_{best}$) is not the same as the estimate that provides the target value($Q_k$). Hence, the effect of coupling further reduces through our update and learning becomes more robust. This is a second advantage of our framework and we believe that this allows for better learning from each sample in the replay memory leading to better understanding on the world. Additionally, we would like to note that the framework is expected to get better at reducing overestimations with increasing number of heads since the chances of a given head providing the greedy action to its own target estimate reduces and so the coupling effect also reduces.\\
\\
We have summarised the Shared Learning framework for any deep $Q$-ensemble algorithm X in Algorithm \ref{algo:slp}.\\
\\
Through the combined ability of transferring knowledge about rewarding states and robust learning updates, Shared Learning is able to replay rewarding sequences more often and also have a better understanding of rewarding states. Empirically, this can be observed from the results on the MDP chain environment in Table \ref{table:mdpreach} and Figure \ref{fig:mdpreach} where the Shared Learning  algorithm is able to make more goal-state visitations than Bootstrap algorithm. Additionally, we can also see the effect of the choice of ``best" head as Shared Learning is able to get to the goal state more often than a Random Head algorithm, where a ``random" head is chosen to give the target estimate. In some sense, Shared Learning can be assumed to have developed a biased implicit curriculum towards rewarding states as simpler goals can be learnt much faster through our framework allowing the agent to focus on more complex goals in the environment.
\begin{table}[h!]
\centering
\begin{tabular}{ |p{1.2cm}|p{1.2cm}|p{1.2cm}|p{1.2cm}|p{1.43cm}|}
 \hline
 No. of states in MDP chain & No. of episodes per run & Bootstrap Algorithm & Random Head Algorithm & Shared Learning Algorithm\\
 \hline
 40 & 150& 56.62& 58.28& \textbf{60.36}\\
 50 & 300& 74.22& 89.96& \textbf{93.18}\\
 60 & 700& 146.48& 166.88& \textbf{184.46}\\
 70 & 1500& 99.96& 151.6& \textbf{164.78}\\
 \hline
\end{tabular}
\caption{Comparison of the number of visitations of the goal state made by each algorithm on an $n$-state MDP chain. The results have been averaged over 50 runs.}
\label{table:mdpreach}
\end{table}
\begin{figure*}[t!]
	\centering
    \begin{subfigure}[b]{0.33\textwidth}
        \includegraphics[width=\textwidth]{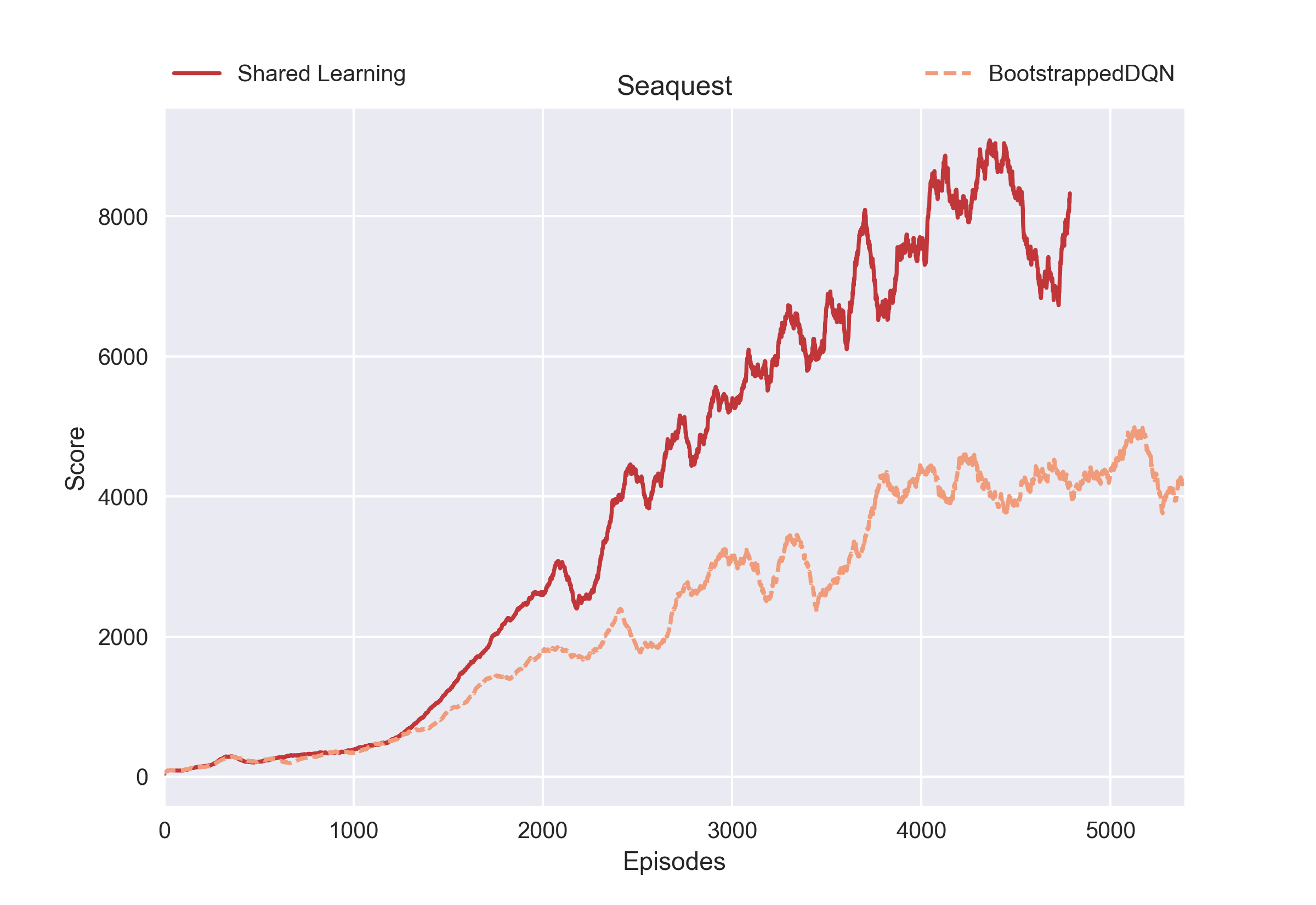}
        \caption{Seaquest}
        \label{fig:Seaquest}
    \end{subfigure}
    \begin{subfigure}[b]{0.33\textwidth}
        \includegraphics[width=\textwidth]{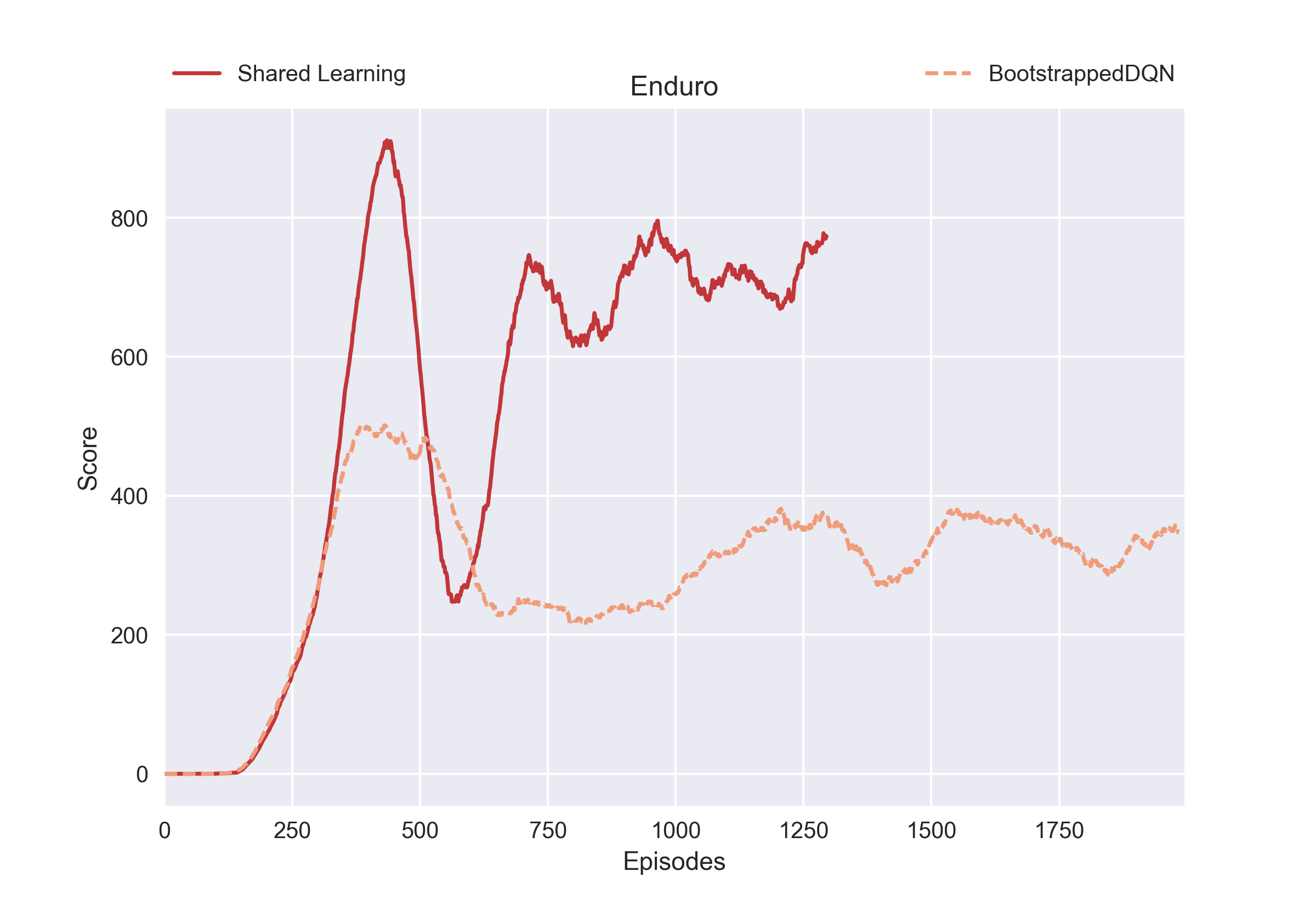}
        \caption{Enduro}
        \label{fig:Enduro}
    \end{subfigure}
    \begin{subfigure}[b]{0.33\textwidth}
        \includegraphics[width=\textwidth]{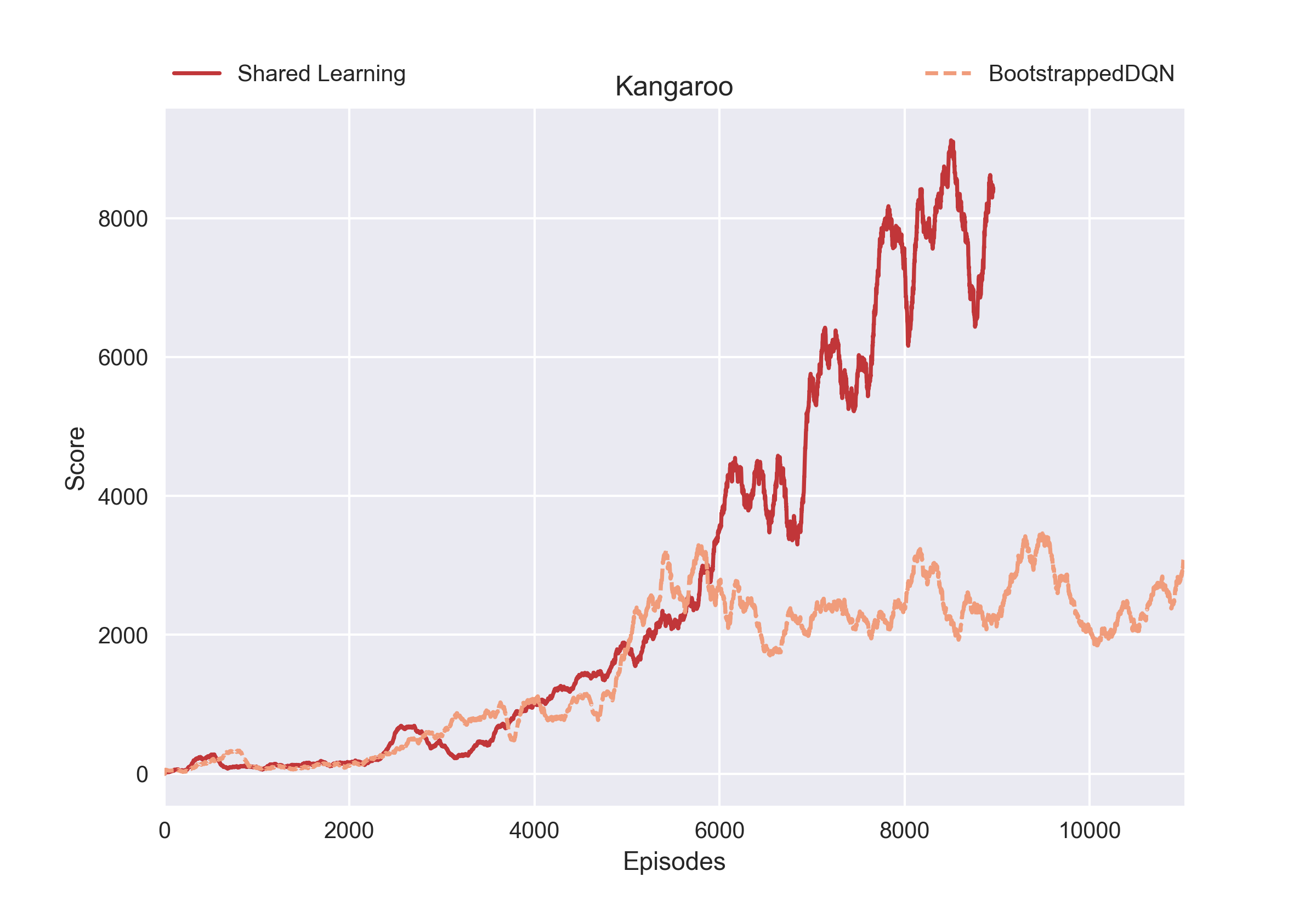}
        \caption{Kangaroo}
        \label{fig:Kangaroo}
    \end{subfigure}
    \begin{subfigure}[b]{0.33\textwidth}
        \includegraphics[width=\textwidth]{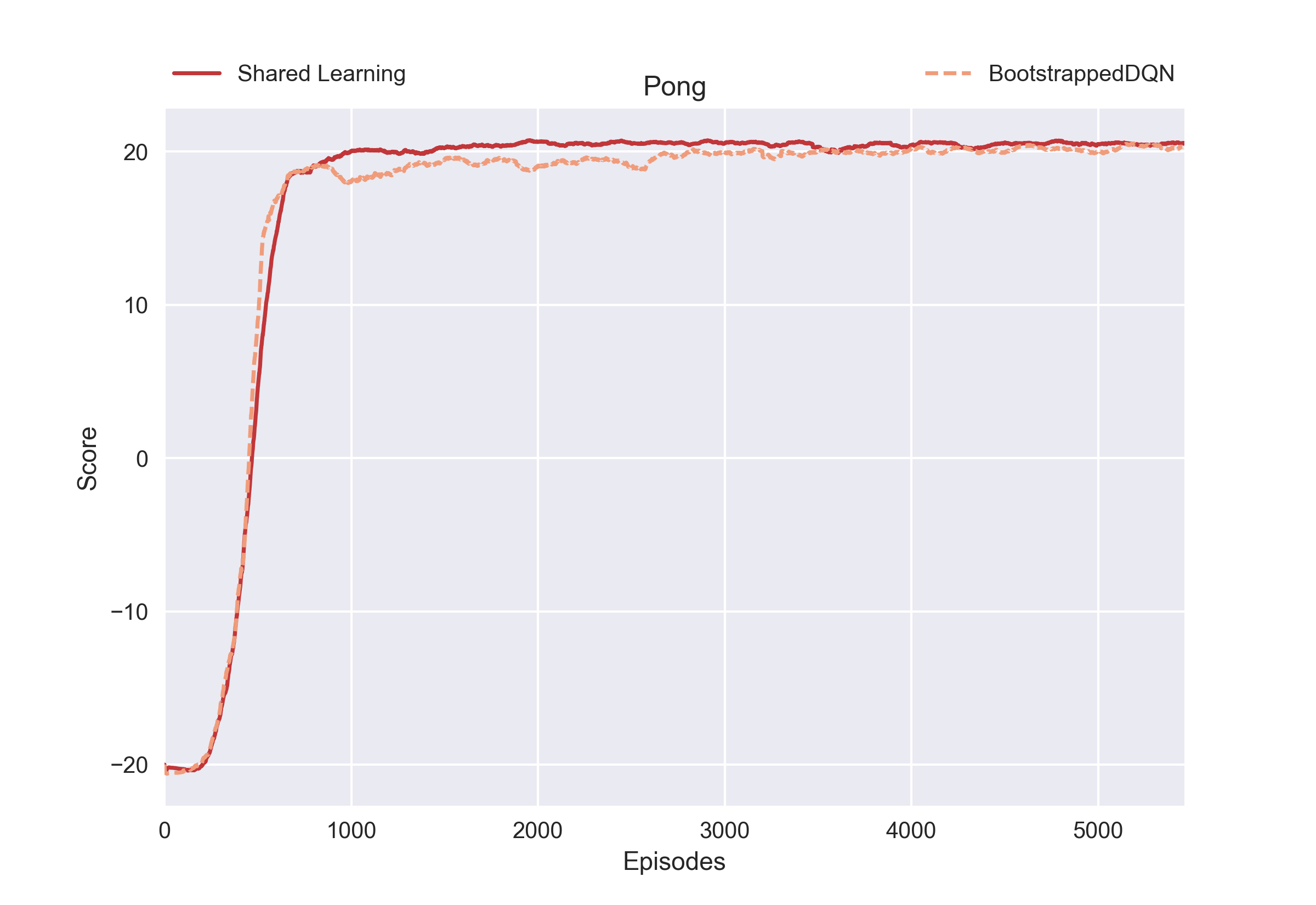}
        \caption{Pong}
        \label{fig:Pong}
    \end{subfigure}
    \begin{subfigure}[b]{0.33\textwidth}
        \includegraphics[width=\textwidth]{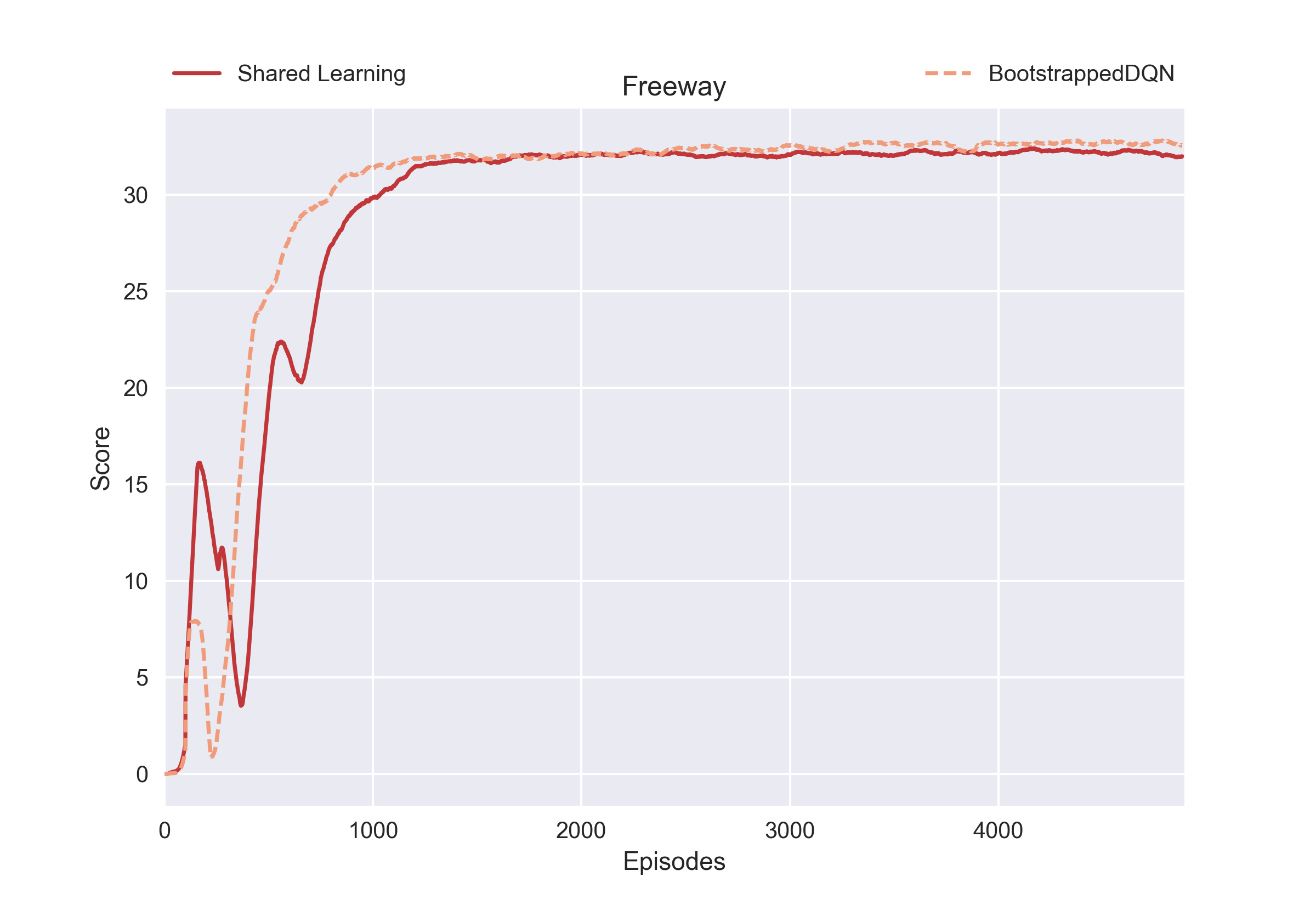}
        \caption{Freeway}
        \label{fig:Freeway}
    \end{subfigure}
    \begin{subfigure}[b]{0.33\textwidth}
        \includegraphics[width=\textwidth]{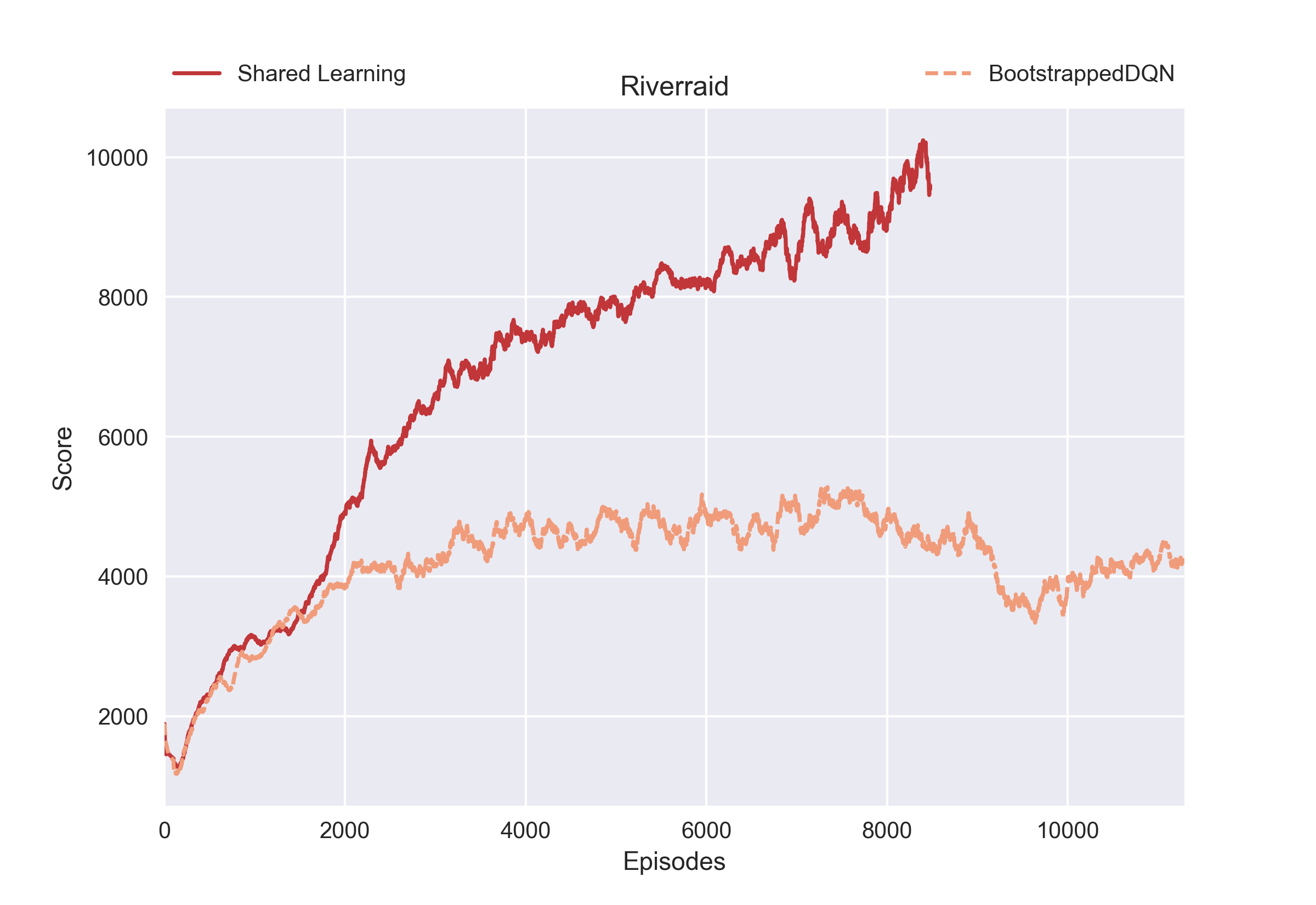}
        \caption{Riverraid}
        \label{fig:Riverraid}
    \end{subfigure}
    \caption{Comparison of Shared Learning-Bootstrap vs BootstrappedDQN algorithm}
   	\label{fig:slbdqn}
\end{figure*}
\begin{figure*}[h!]
	\centering
    \begin{subfigure}[b]{0.33\textwidth}
        \includegraphics[width=\textwidth]{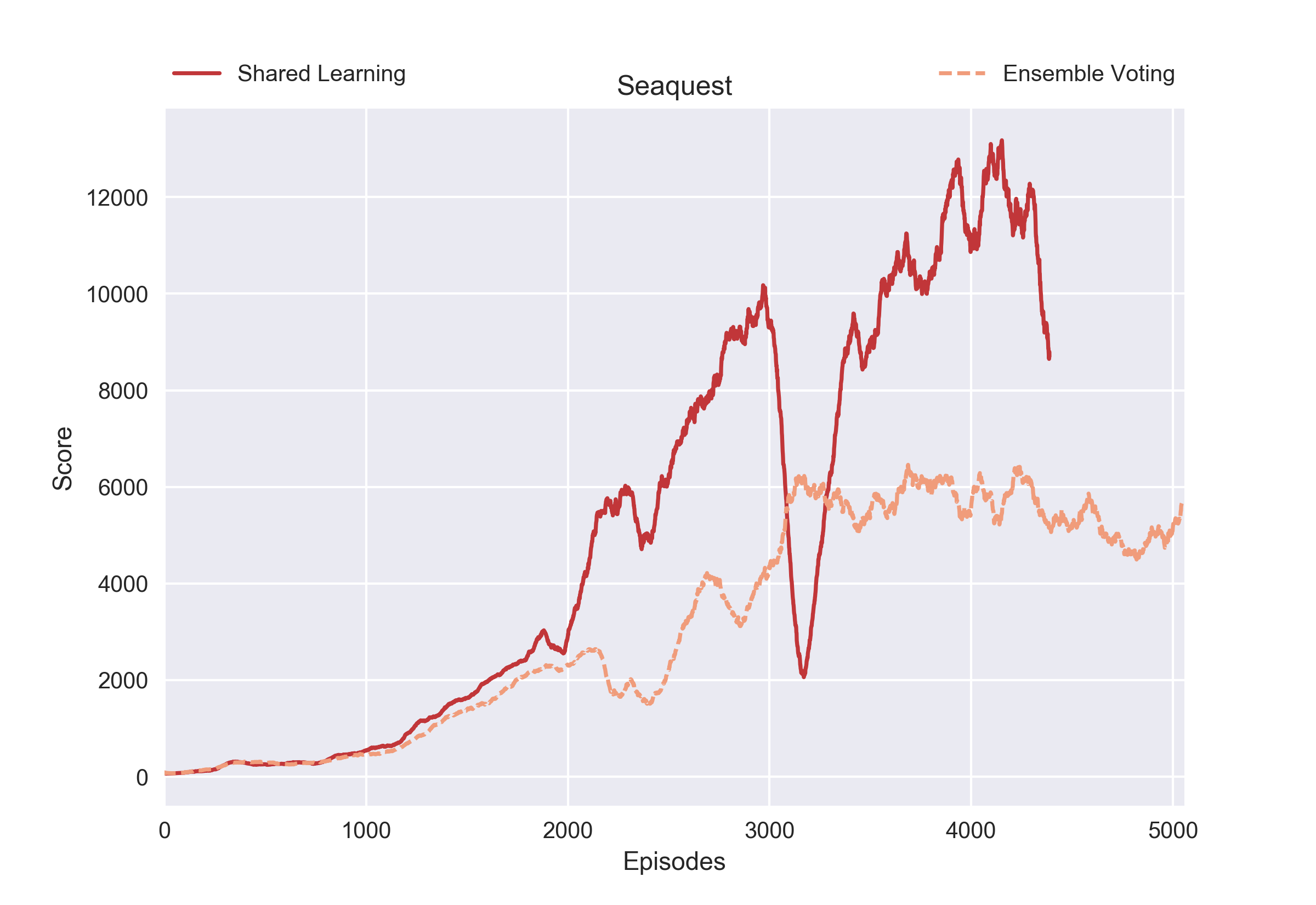}
        \caption{Seaquest}
        \label{fig:Seaquest_EV}
    \end{subfigure}
    \begin{subfigure}[b]{0.33\textwidth}
        \includegraphics[width=\textwidth]{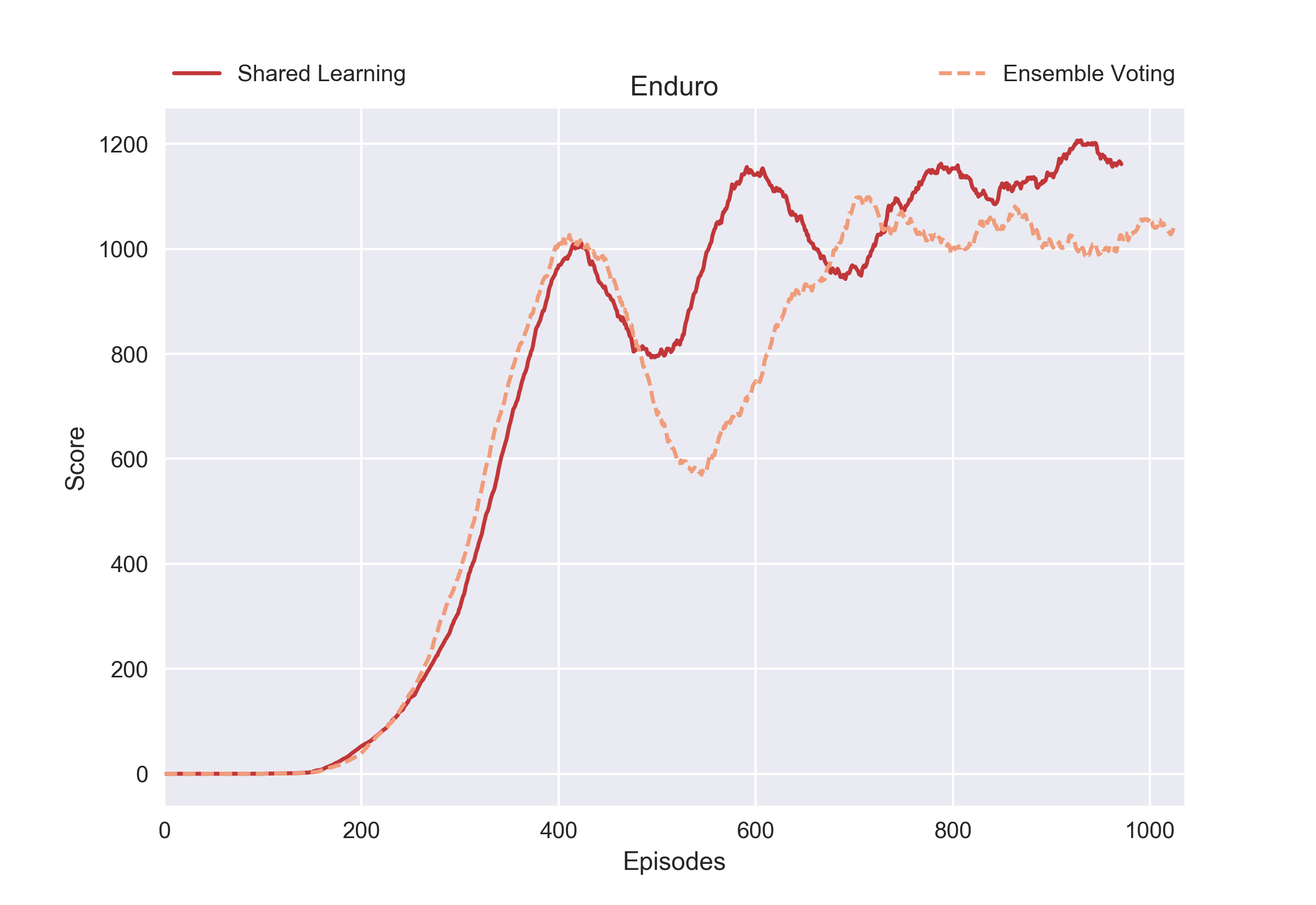}
        \caption{Enduro}
        \label{fig:Enduro_EV}
    \end{subfigure}
    \begin{subfigure}[b]{0.33\textwidth}
        \includegraphics[width=\textwidth]{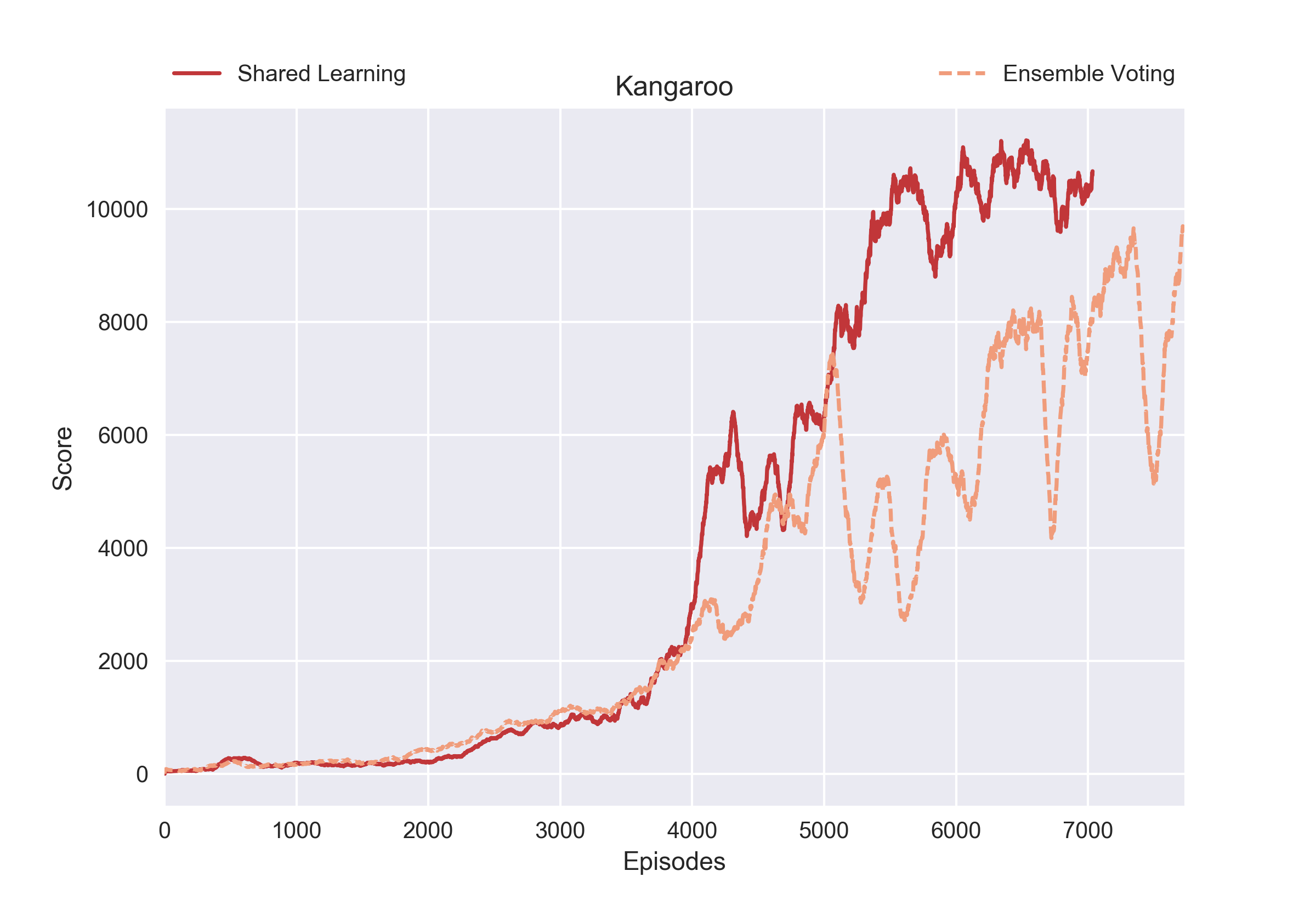}
        \caption{Kangaroo}
        \label{fig:Kangaroo_EV}
    \end{subfigure}
    \begin{subfigure}[b]{0.33\textwidth}
        \includegraphics[width=\textwidth]{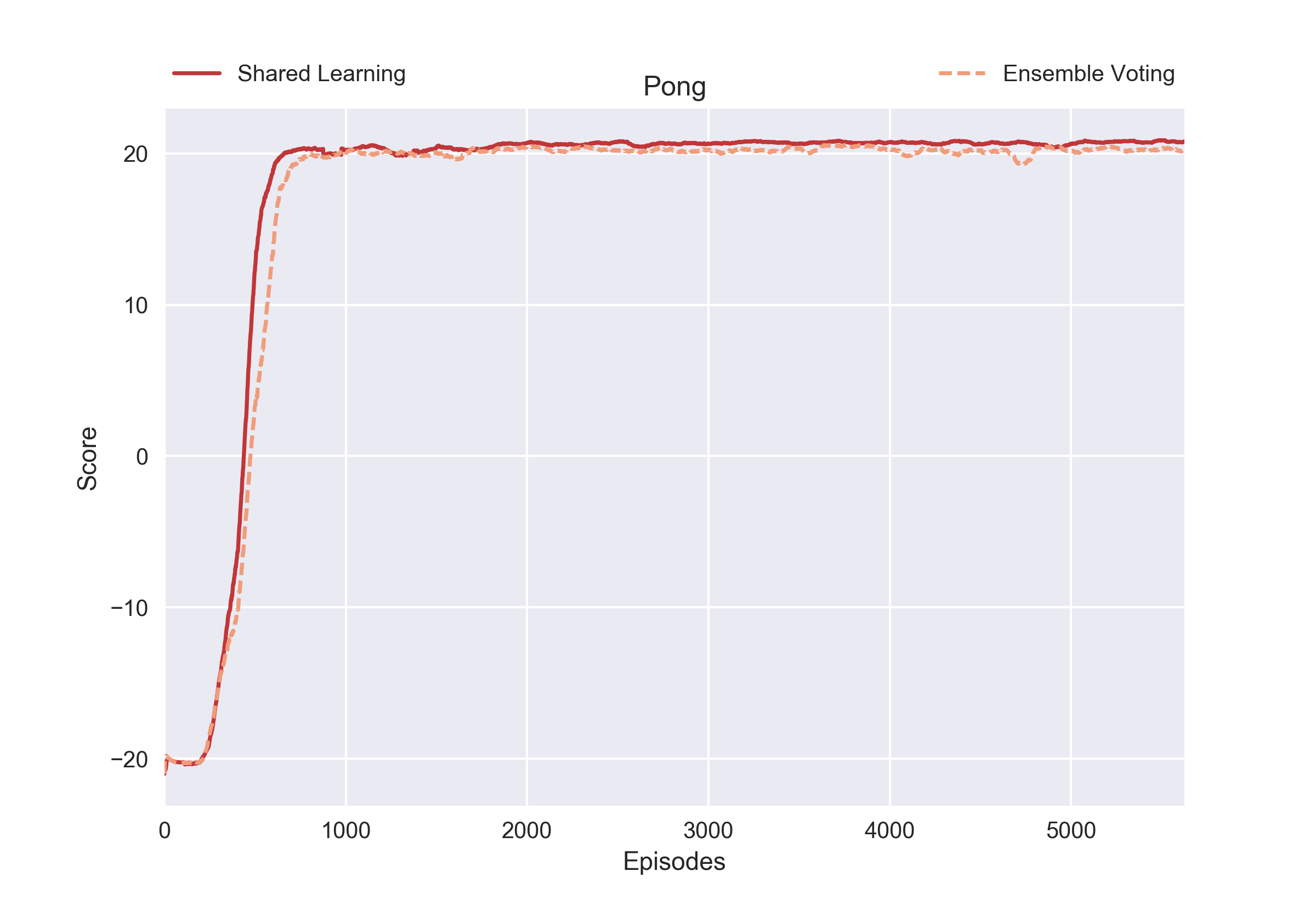}
        \caption{Pong}
        \label{fig:Pong_EV}
    \end{subfigure}
    \begin{subfigure}[b]{0.33\textwidth}
        \includegraphics[width=\textwidth]{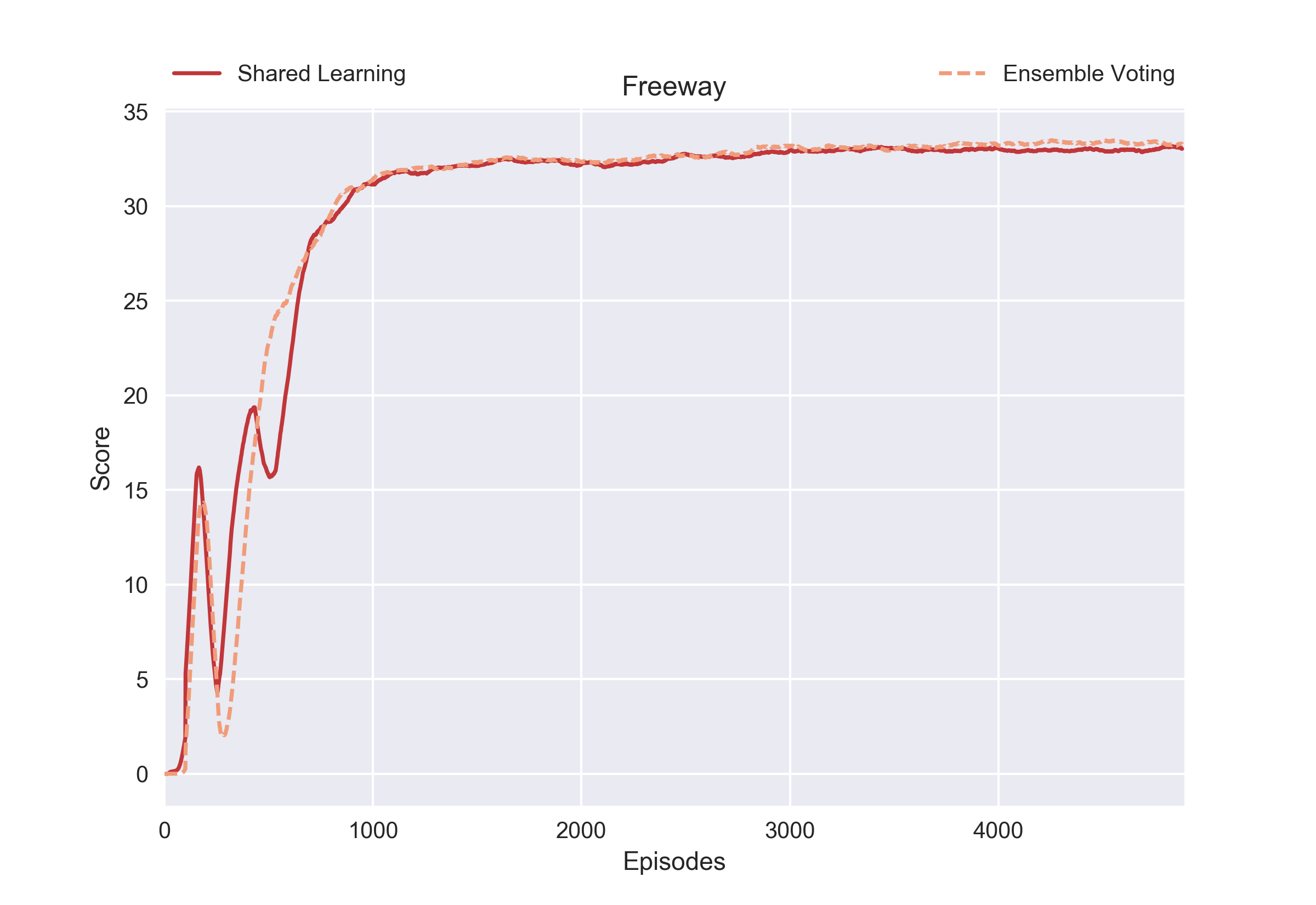}
        \caption{Freeway}
        \label{fig:Freeway_EV}
    \end{subfigure}
    \begin{subfigure}[b]{0.33\textwidth}
        \includegraphics[width=\textwidth]{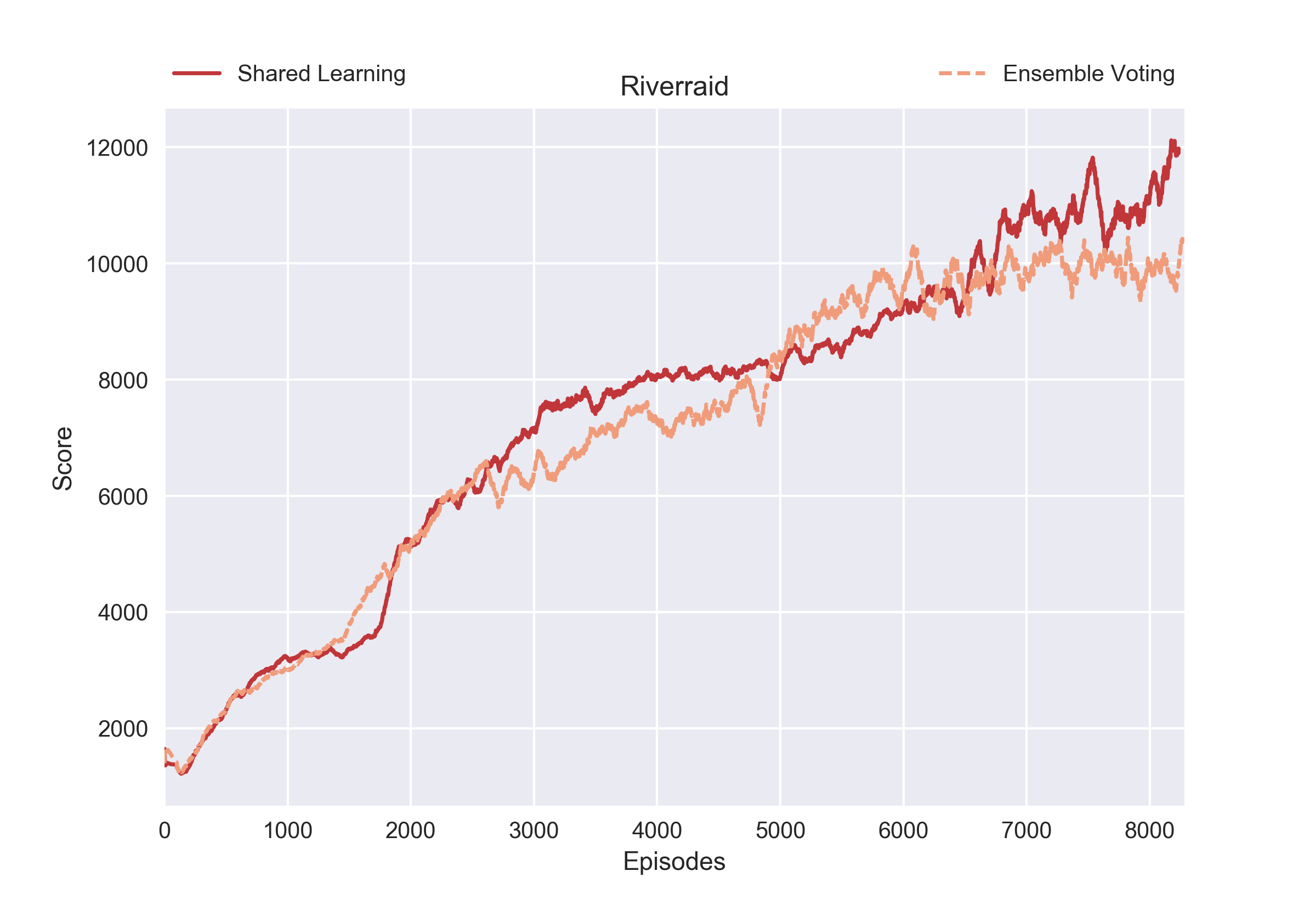}
        \caption{Riverraid}
        \label{fig:Riverraid_EV}
    \end{subfigure}
    \caption{Comparison of Shared Learning-Ensemble Voting vs Ensemble Voting algorithm}
    \label{fig:slev}
\end{figure*}
\begin{algorithm}[h!]
\caption{Shared Learning - X}
\label{algo:slp}
\begin{algorithmic}[1]
\footnotesize 
\INPUT Value function networks $Q$ with $K$ outputs $\{Q_k\}_{k=1}^K$
\State Let B be a replay buffer storing experience for training and select\_best\_int be the interval during which the best head is selected.\;
\State numSteps $\leftarrow$ 0\;
\State best\_head $ \sim$ Uniform$\{1,2,\dots,K\}$\;
 \For {each episode}
  \State Obtain initial state from environment $s_0$\;
  \For {step $t=1,\dots$ until end of episode}
  \State Pick an action according to $Q$-ensemble algorithm X\;
  \State Take action $a_t$ and receive state $s_{t+1}$ and reward $r_t$ 
  \State Store transition ($s_t$,$a_t$,$r_t$,$s_{t+1}$) in B
  \State Sample minibatch from B
  \State Train network with loss function given in equation \ref{eq:slp}\;
  \State numSteps $\leftarrow$ numSteps + 1\;
  \If{numSteps $mod$ select\_best\_int $== 0$}
   \State best\_head = argmax$_k Q_k(s_t,a_t)$
   \EndIf
  \EndFor
 \EndFor
\end{algorithmic}
\end{algorithm}
\section{Experiments and Discussion}
In this section, we try to answer the following questions:
\begin{itemize}
\item Can Shared Learning become data-efficient with deep reinforcement learning algorithms?
\item Can the algorithm be extended to exploitation $Q$-ensemble algorithms?
\end{itemize}
To this end, we test the efficacy of the Shared Learning framework on an exploration $Q$-ensemble algorithm and an exploitation $Q$-ensemble algorithm with BootstrappedDQN and Ensemble Voting \cite{2017arXiv170601502C} respectively. The hyperparameters were tuned for 6 Atari Games (Seaquest, Riverraid, Kangaroo, Pong, Freeway and Enduro) and were kept constant throughout both experiments.

In our experiments, the most important parameter to be dealt with is the frequency with which we choose the ``best" head among the value function estimates. Here, we have chosen the frequency to be 10,000 steps, i.e., the head that is required to perform knowledge sharing is chosen once every 10,000 steps. The DQN code has been taken from OpenAI baselines\footnote{\url{https://github.com/openai/baselines}} and our algorithms are evaluated on the Atari Games in OpenAI Gym \cite{2016arXiv160601540B}, which is simulated by Arcade Learning Environment \cite{bellemare2013arcade} and trained for 40 million frames on each game. For the graphs, we have plotted the learning curves for the algorithms against the number of episodes played over 40 million frames. We believe that in doing so we get to see that our proposed framework plays lesser number of episodes within the same number of timesteps and still gets to a higher score which essentially shows that the framework makes for data efficiency.
\subsection{Shared Learning-BootstrappedDQN}
The training curves for Shared Learning-BootstrappedDQN vs BootstrappedDQN is shown in Figure \ref{fig:slbdqn}. We have also summarised the highest scores obtained during training on the different games in Table \ref{table:slbscores}. We refer to the algorithm BootstrappedDQN as BDQN and the application of Shared Learning on BDQN as SLBDQN. From the graph and the table, we can see that Shared Learning is able to do equal or better than BootstrappedDQN on 5 out of 6 games in terms of score and it is also able to achieve these scores with lesser data. We claim this is because of the ability of the framework to replay rewarding trajectories much more often because of the knowledge sharing that happens between the value function estimates making the framework data-efficient for the exploration algorithm. Further, we see that the graphs have an increasing slope in Seaquest, Kangaroo and Enduro which indicates that we can get better scores with more training.
\begin{table}[h!]
\centering
\begin{tabular}{ |p{1.2cm}|p{1.4cm}|p{1.5cm}|p{1.1cm}|p{1.3cm}|  }
 \hline
 Game & BDQN-MeanMax & SLBDQN-MeanMax & BDQN-EpMax  & SLBDQN-EpMax\\
 \hline
 Pong & 20.53& \textbf{20.74}& \textbf{21}& \textbf{21}\\
 Kangaroo & 3476& \textbf{9120}& 10600& \textbf{15000}\\
 Riverraid & 5307.9& \textbf{10242.8}& 9280& \textbf{15370}\\
 Enduro & 501.48& \textbf{911.47}& 1007& \textbf{1869}\\
 Freeway & \textbf{32.81}& 32.39& \textbf{34}& 33\\
 Seaquest & 4989.4& \textbf{9086.4}& 14880& \textbf{23970}\\
 \hline
\end{tabular}
\caption{Experimental results for Shared Learning-Bootstrap and BootstrappedDQN experiments. Here MeanMax represents the scores from the mean curves in Figure \ref{fig:slbdqn}, EpMax represents the maximum score achieved in an episode by the algorithm.}
\label{table:slbscores}
\end{table}
\vspace{-0.5cm}
\subsection{Shared Learning-Ensemble Voting}
The training curves for Shared Learning-Ensemble Voting vs Ensemble Voting is shown in Figure \ref{fig:slev}. We have also summarised the highest scores obtained during training on the different games in Table \ref{table:slevscores}. We refer to the algorithm Ensemble Voting as EV and the application of Shared Learning on EV as SLEV. From the graph and the table, we can see that Shared Learning is able to do equal or better than Ensemble Voting on 5 out of 6 games in terms of score and it is also able to achieve these scores with lesser data due to the same reasons as mentioned in the previous subsection. However, we believe that the data efficiency and the improved scores are not as profound as in BootstrappedDQN because we are applying a biased curriculum on an exploitation algorithm.
\begin{table}[h!]
\centering
\begin{tabular}{ |p{1.2cm}|p{1.4cm}|p{1.5cm}|p{1.1cm}|p{1.3cm}|  }
 \hline
 Game & EV-MeanMax & SLEV-MeanMax & EV-EpMax  & SLEV-EpMax\\
 \hline
 Pong & 20.6& \textbf{20.87}& \textbf{21}& \textbf{21}\\
 Kangaroo & 9694& \textbf{11215}& 14700& \textbf{15200}\\
 Riverraid & 10443.1& \textbf{12118.2}& 15770& \textbf{16530}\\
 Enduro & 1098.5& \textbf{1206.77}& 1882& \textbf{1940}\\
 Freeway & \textbf{33.48}& 33.16& \textbf{34}& \textbf{34}\\
 Seaquest & 6456.6& \textbf{13176}& 20360& \textbf{35180}\\
 \hline
\end{tabular}
\caption{Experimental results for Shared Learning-Ensemble Voting and Ensemble Voting experiments. Here MeanMax represents the scores from the mean curves in Figure \ref{fig:slev}, EpMax represents the maximum score achieved in an episode by the algorithm.}
\label{table:slevscores}
\end{table}
\vspace{-0.2cm}
\section{Conclusion \& Future Work}
We have proposed a low computational overhead framework Shared Learning to make $Q$-ensemble algorithms data-efficient. We have also shown that the framework is applicable to exploration and exploitation algorithms such as BootstrappedDQN and Ensemble Voting respectively. Our action advice in learning update framework shows a significant improvements in game score for some of the Atari Games indicating faster learning through robust estimates.\\
\begin{figure}[h!]
\centering
\includegraphics[width=0.44\textwidth]{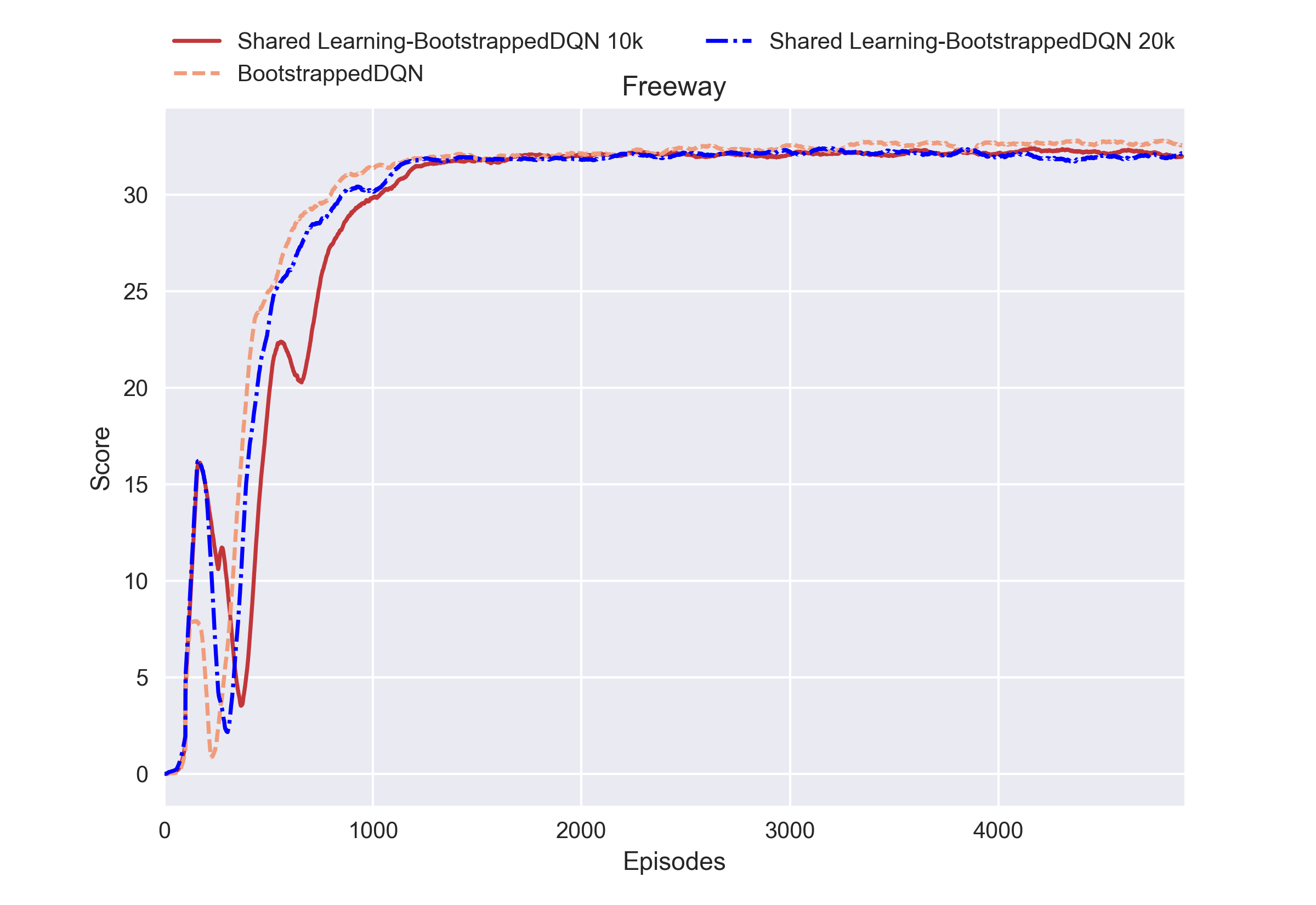}
\caption{Freeway using 20,000-step interval for choosing the ``best head", is able to learn much faster initially than the 10,000-step interval. Analysis done for SLBDQN vs BDQN.}
\label{fig:freeway_80k}
\end{figure}
In this paper, we have fixed the interval for choosing the ``best" head to share knowledge among the other heads. However, we could look to tune this hyperparameter and make it dynamic. For example, we can see in Figure \ref{fig:freeway_80k} that upon changing the parameter from 10,000 steps to 20,000 steps for Freeway, we can make the framework learn faster initially. We could also tune the number of heads to share knowledge from to improve the robustness of the learning rule. Since, we have shown that our framework is able to perform better learning updates when compared to DoubleDQN, we could test the effect of our learning update on masking BootstrappedDQN. We leave these modifications for Shared Learning to future work.
\section{Acknowledgements}
We would like to thank Manu Srinath Halvagal, EPFL, for some useful discussions as well Joe Eappen, IIT Madras, and Yash Chandak, UMass Amherst, for their valuable feedback on the work.
\bibliographystyle{aaai}
\bibliography{ref}

\end{document}